\documentclass{article}

\usepackage{arxiv}
\usepackage{float}
\usepackage[T1]{fontenc}    
\usepackage[hidelinks]{hyperref}       
\usepackage{url}            
\usepackage{booktabs}       
\usepackage{amsfonts}       
\usepackage{nicefrac}       
\usepackage{microtype}      
\usepackage{graphicx}

\usepackage[T1]{fontenc}

\usepackage{longtable}

\usepackage{xcolor}
\usepackage{listings}

\colorlet{punct}{red!60!black}
\definecolor{background}{HTML}{EEEEEE}
\definecolor{delim}{RGB}{20,105,176}
\colorlet{numb}{magenta!60!black}

\usepackage{adjustbox}
\usepackage{makecell}
\usepackage{amsmath}

\lstdefinelanguage{json}{
	basicstyle=\normalfont\ttfamily,
	numbers=left,
	numberstyle=\scriptsize,
	stepnumber=1,
	numbersep=8pt,
	showstringspaces=false,
	breaklines=true,
	frame=lines,
	backgroundcolor=\color{background},
	literate=
	*{0}{{{\color{numb}0}}}{1}
	{1}{{{\color{numb}1}}}{1}
	{2}{{{\color{numb}2}}}{1}
	{3}{{{\color{numb}3}}}{1}
	{4}{{{\color{numb}4}}}{1}
	{5}{{{\color{numb}5}}}{1}
	{6}{{{\color{numb}6}}}{1}
	{7}{{{\color{numb}7}}}{1}
	{8}{{{\color{numb}8}}}{1}
	{9}{{{\color{numb}9}}}{1}
	{:}{{{\color{punct}{:}}}}{1}
	{,}{{{\color{punct}{,}}}}{1}
	{\{}{{{\color{delim}{\{}}}}{1}
	{\}}{{{\color{delim}{\}}}}}{1}
	{[}{{{\color{delim}{[}}}}{1}
	{]}{{{\color{delim}{]}}}}{1},
}

\title{BanglaWriting: A multi-purpose offline Bangla handwriting dataset}

\date{} 					

\author{
	\href{https://orcid.org/0000-0001-5738-1631}{\includegraphics[scale=0.06]{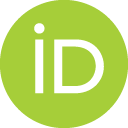}\hspace{1mm}M. F. Mridha} \\
	Department of Computer Science \& Engineering\\
	Bangladesh University of Business \& Technology\\ 
	Dhaka, Bangladesh \\
	\texttt{firoz@bubt.edu.bd} \\
	\And
	\href{https://orcid.org/0000-0001-7375-9040}{\includegraphics[scale=0.06]{orcid.png}\hspace{1mm}Abu Quwsar Ohi} \\
	Department of Computer Science \& Engineering\\
	Bangladesh University of Business \& Technology\\ 
	Dhaka, Bangladesh \\
	\texttt{quwsarohi@gmail.com} \\
	\And
	M. Ameer Ali \\
	Department of Information Technology\\
	Department of Computer Science \& Engineering\\
	Bangladesh University of Business \& Technology\\ 
	Dhaka, Bangladesh \\
	\texttt{dmaa730@gmail.com} \\
	\And
	Mazedul Islam Emon \\
	Department of Information Technology\\
	Department of Computer Science \& Engineering\\
	Bangladesh University of Business \& Technology\\ 
	Dhaka, Bangladesh \\
	\texttt{emon.bubt3382@gmail.com} \\
	\And
	Muhammad Mohsin Kabir \\
	Department of Information Technology\\
	Department of Computer Science \& Engineering\\
	Bangladesh University of Business \& Technology\\ 
	Dhaka, Bangladesh \\
	\texttt{m97kabir2@gmail.com} \\
}

\renewcommand{\headeright}{Dataset Manual}
\renewcommand{\undertitle}{Dataset Manual \href{https://data.mendeley.com/datasets/r43wkvdk4w/1}{[Link]}}

\begin{document}
\maketitle

\begin{abstract}
	This article presents a Bangla handwriting dataset named BanglaWriting that contains single-page handwritings of 260 individuals of different personalities and ages. Each page includes bounding-boxes that bounds each word, along with the unicode representation of the writing. This dataset contains 21,234 words and 32,787 characters in total. Moreover, this dataset includes 5,470 unique words of Bangla vocabulary. Apart from the usual words, the dataset comprises 261 comprehensible overwriting and 450 handwritten strikes and mistakes. All of the bounding-boxes and word labels are manually-generated. The dataset can be used for complex optical character/word recognition, writer identification, handwritten word segmentation, and word generation. Furthermore, this dataset is suitable for extracting age-based and gender-based variation of handwriting.
	
\end{abstract}


\section*{Specifications Table} 
{\fontsize{7.5pt}{9pt}\selectfont
	\begin{longtable}{|p{33mm}|p{94mm}|}
		\hline
		\endhead
		\hline
		\endfoot
		Subject                & Computer Vision and Pattern Recognition\\
		\hline                         
		Specific subject area  & Optical character recognition, word segmentation, writer identification\\
		\hline
		Type of data           & Image\newline
		JSON\\
		\hline
		
		How data were acquired & The images of the handwriting were captured using scanners and smartphone cameras. Each of the handwriting-images was cropped and annotated manually. \\
		
		\hline                         
		Data format            & Raw data\newline
		Converted data\newline
		Annotations\\
		\hline                         
		Parameters for         
		data\newline 
		collection             & Scanner: HP Scanjet 2400\newline
		Smartphone camera: Xiaomi Redmi 6, Xiaomi Redmi 7 \newline
		A single image contains the handwriting of an individual. Each individual is identified using age, gender, and unique person id. The handwritten words are segmented using bounding-boxes. Each of the bounding-boxes contains the characters that are written. Labelme \cite{labelme2016} software is used to draw and label the bounding-boxes.\\  
		
		\hline
		Description of          
		data\newline 
		collection             & The writings were conducted using regular stationery products. Writers were advised to write on a random topic. Only one page of writing was collected from each individual. The handwritings were further captured using scanners and smartphone cameras. Each captured image was cropped and annotated manually. \\
		\hline                         
		Data source location   & Institution: Bangladesh University of Business \& Technology\newline
		District: Dhaka, Kishoreganj, Gopalganj, Comilla, Gazipur, Tangail, Netrakona, Mymensingh \newline
		Country: Bangladesh\\
		\hline                         
		\hypertarget{target1}
		{Data accessibility}   & Repository name: Mendeley\newline
		Data identification number: 10.17632/r43wkvdk4w.1\newline
		Direct URL to data: \href{https://data.mendeley.com/datasets/r43wkvdk4w/1}{https://data.mendeley.com/datasets/r43wkvdk4w/1}\\                         
	\end{longtable}
}

\setcounter{table}{0}

\section*{Value of the Data}

\begin{itemize}
	\item The dataset exploits possibilities and usage of handwritings from scanned and pictured documents. The usage of scanned and pictured forms in the recognition and identification process is often termed as an offline approach.
	
	\item The dataset is suitable for machine learning \cite{michie1994machine} models, deep learning \cite{lecun2015deep} models, producing embedding vectors \cite{ohi2020autoembedder} of handwriting, etc.
	
	\item The dataset exploits all possible potentials of Bangla handwriting \cite{marti2002iam}. The dataset contains bounding-box annotations for each handwritten word, unicode representation for each written word, and writer information for each document. Therefore, the dataset is suitable for word segmentation, optical character recognition, writer identification, writer verification, and handwriting generation.
	
	\item The dataset contains raw images (without any pre-processing) of each document. The dataset also contains supplementary pre-processing scripts to suspend excess lighting and noises.

	\item The dataset can be used to explore writing patterns related to age and gender.
	
\end{itemize}

\section{Data Description}

\begin{figure}[!h]
	\centering
	\includegraphics[width=0.9\linewidth]{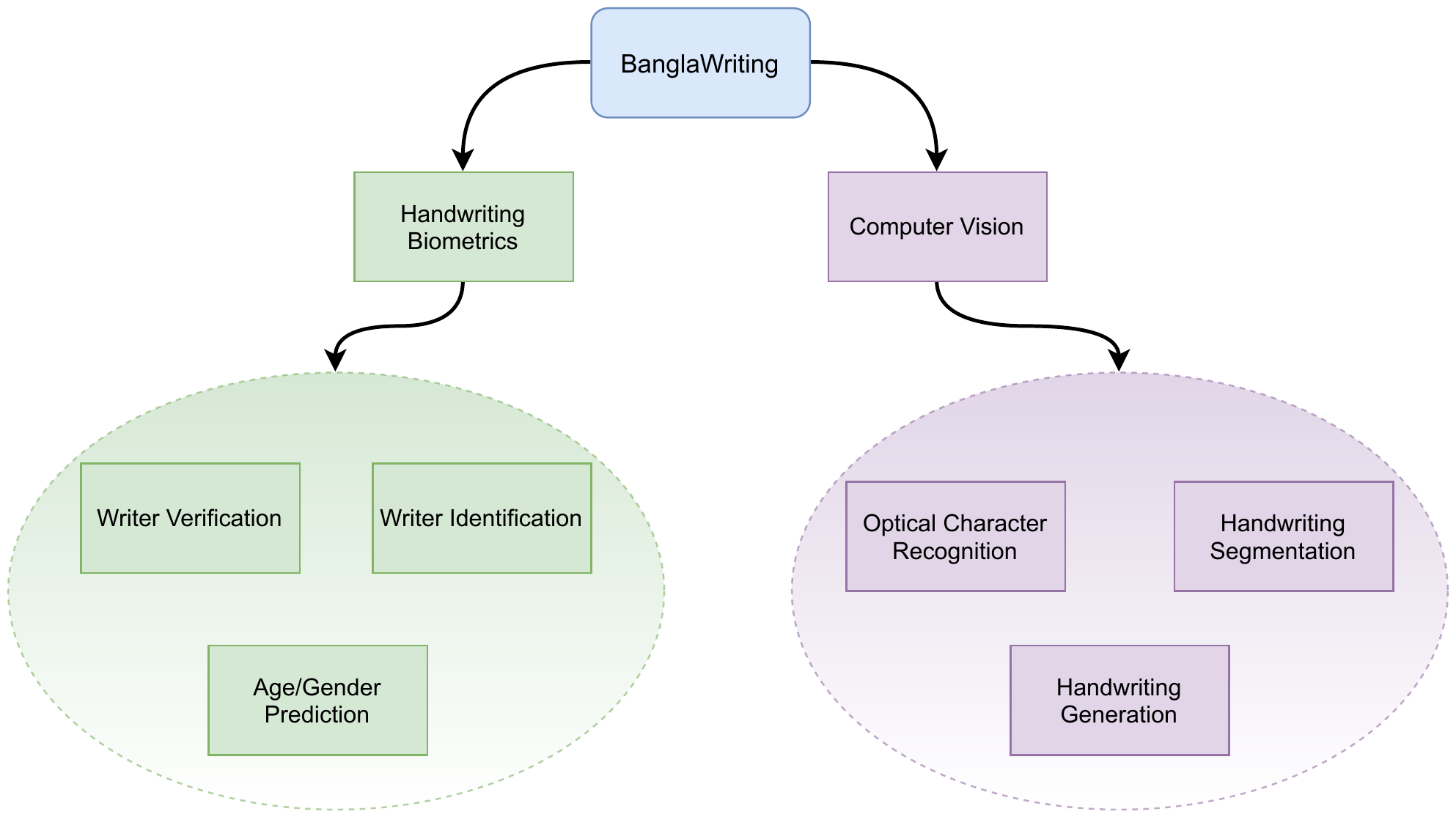}
	\caption{The BanglaWriting dataset can be used for handwriting biometrics and computer vision-specific tasks. The dataset has possibilities in various fields, including identifying writers, to generating handwritings from unicode.}
	\label{fig:applications}
\end{figure}

BanglaWriting, the dataset presented in this paper, aims to provide a preferable handwriting dataset that is enriched from every dimension. The dataset can be used in diverse machine learning and deep learning based applications. It can be implemented in handwriting biometric tasks, including identification, verification, and age/gender estimation. Further, the dataset has possibilities for specific computer vision tasks such as optical character recognition and handwriting segmentation. Moreover, the dataset has the capability of fueling generative handwriting models. Fig. \ref{fig:applications} illustrates the possible domains on which the dataset can contribute.

This dataset's construction and usage are different from usual Bangla datasets \cite{biswas2017banglalekha}. The currently available datasets for Bangla writing only include isolated character writings. Whereas, the BanglaWriting dataset contains word-based writing with bounding-boxes. The dataset is implemented based on well-known offline handwriting, and writer recognition datasets \cite{marti2002iam}. Table \ref{tab:comparison} presents a comparison BanglaWriting dataset with some of the popular datasets of diverse languages. Most of the bigger datasets (such as KHATT \cite{mahmoud2014khatt}, IAM \cite{marti2002iam}) include some automated and pre-estimated parameters to label the data. In comparison, the annotations and labels of the BanglaWriting dataset are manually determined. Hence from the overall evaluation, it can be concluded that the BanglaWriting dataset attains a marginal amount of quality data.

\begin{table}
	\centering
	\caption{The table illustrates a quantitative comparison of the BanglaWriting dataset with some famous datasets in different languages. The BanglaWriting dataset targets almost all possible domains of interest in offline handwriting processing. In general, most datasets neglect various classes (overwriting, random strikes) of handwriting. Hence, we exclude the number of classes in comparison.}
	\label{tab:comparison}
	\begin{adjustbox}{width=\columnwidth,center}
		\begin{tabular}{|c|c|c|c|c|c|c|}
			\hline
			\textbf{Dataset} & \textbf{Language} & \textbf{Writers} & \textbf{Total Documents} & \textbf{Word Count} & \makecell{\textbf{Word-level} \\ \textbf{Bounding-box}} \\
			\hline
			
			RIMES \cite{grosicki12008rimes} & French & 1300 & 12723 & 300000 & Yes \\ \hline		
			
			KHATT \cite{mahmoud2014khatt} & Arabic & 1000 & 2000 & 165890 & No \\ \hline
			
			IAM \cite{marti2002iam} & English & 400 & 1066 & 82227 & Yes \\ \hline
			
			\textbf{BanglaWriting} & \textbf{Bangla} & \textbf{260} & \textbf{260} & \textbf{21234} & \textbf{Yes} \\ \hline
			
			Firemaker \cite{schomaker2000forensic} & Dutch & 252 & 1008 & - & No \\ \hline
			
			AHDB \cite{al2002data} & Arabic & 105 & - & 10000 & Yes \\ \hline
		\end{tabular}
	\end{adjustbox}
\end{table}

\begin{figure}[!h]
	\begin{center}
		\includegraphics[width=0.9\linewidth]{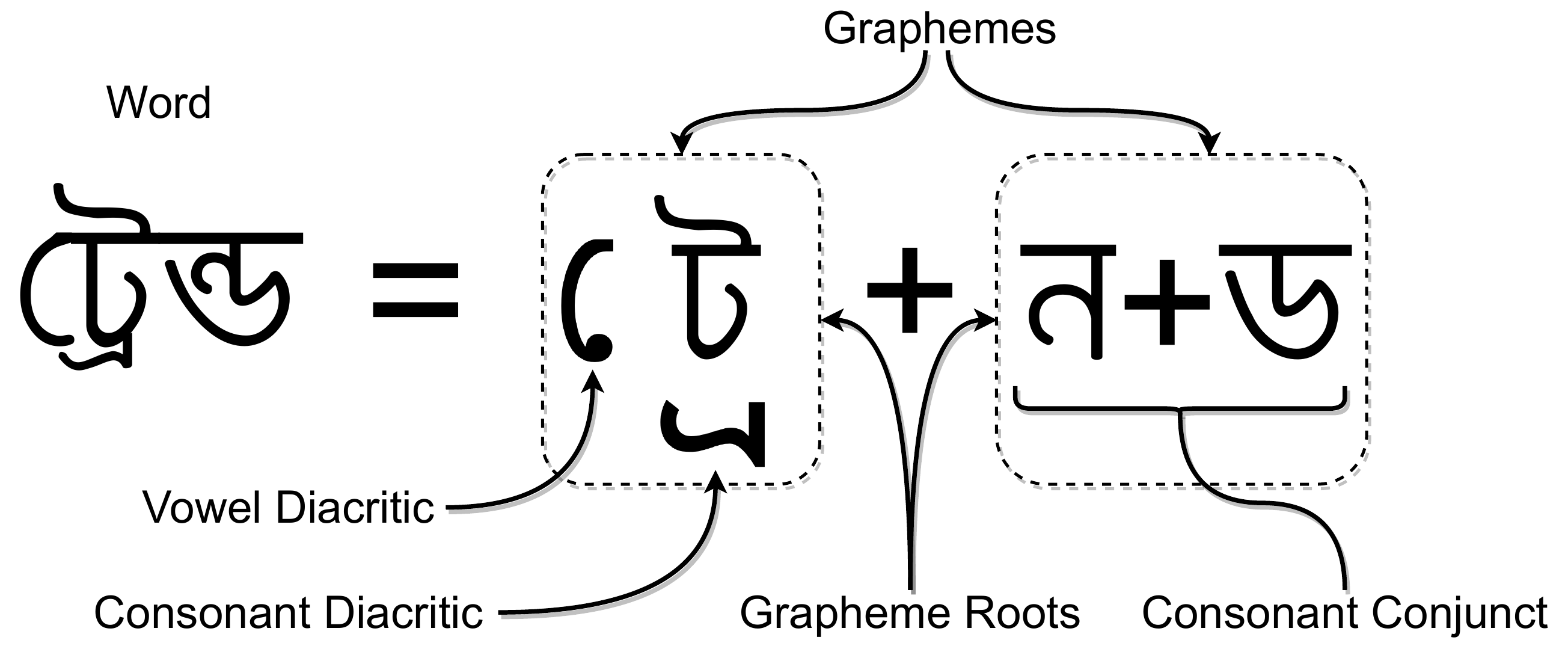}
		\caption{Graphemes are the smallest unit of meaningful writing. A grapheme always contains a grapheme root. In the Bangla writing system, a grapheme may have one vowel and one consonant diacritic. Occasionally, a grapheme may include consonant conjuncts as it's grapheme root. The figure is derived from \cite{banglaAI_2019,alam2021large}.}
		\label{fig:construction}
	\end{center}
\end{figure}

\begin{figure}[!h]
	\begin{center}
		\includegraphics[width=0.49\linewidth]{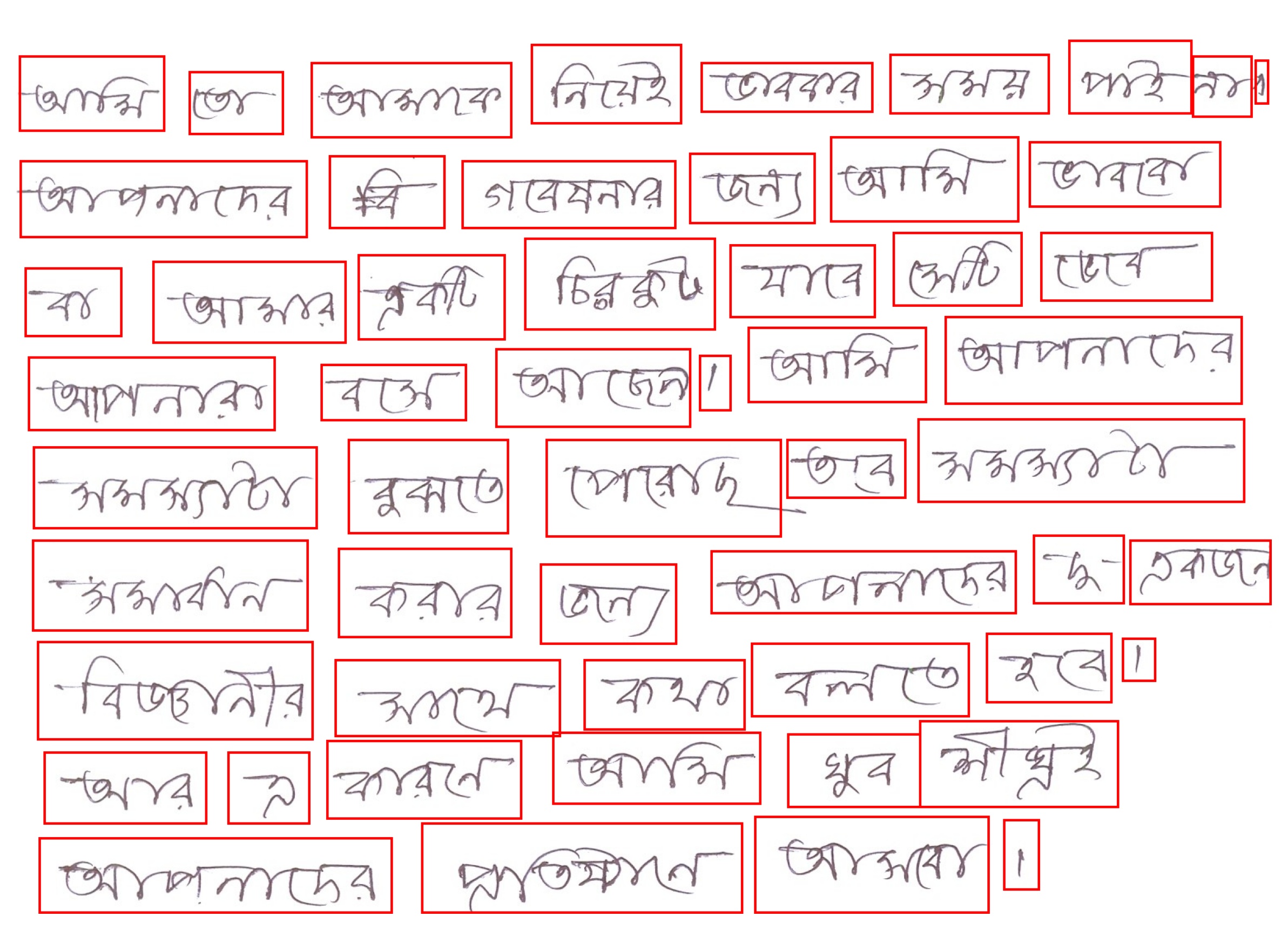}
		\includegraphics[scale=0.35]{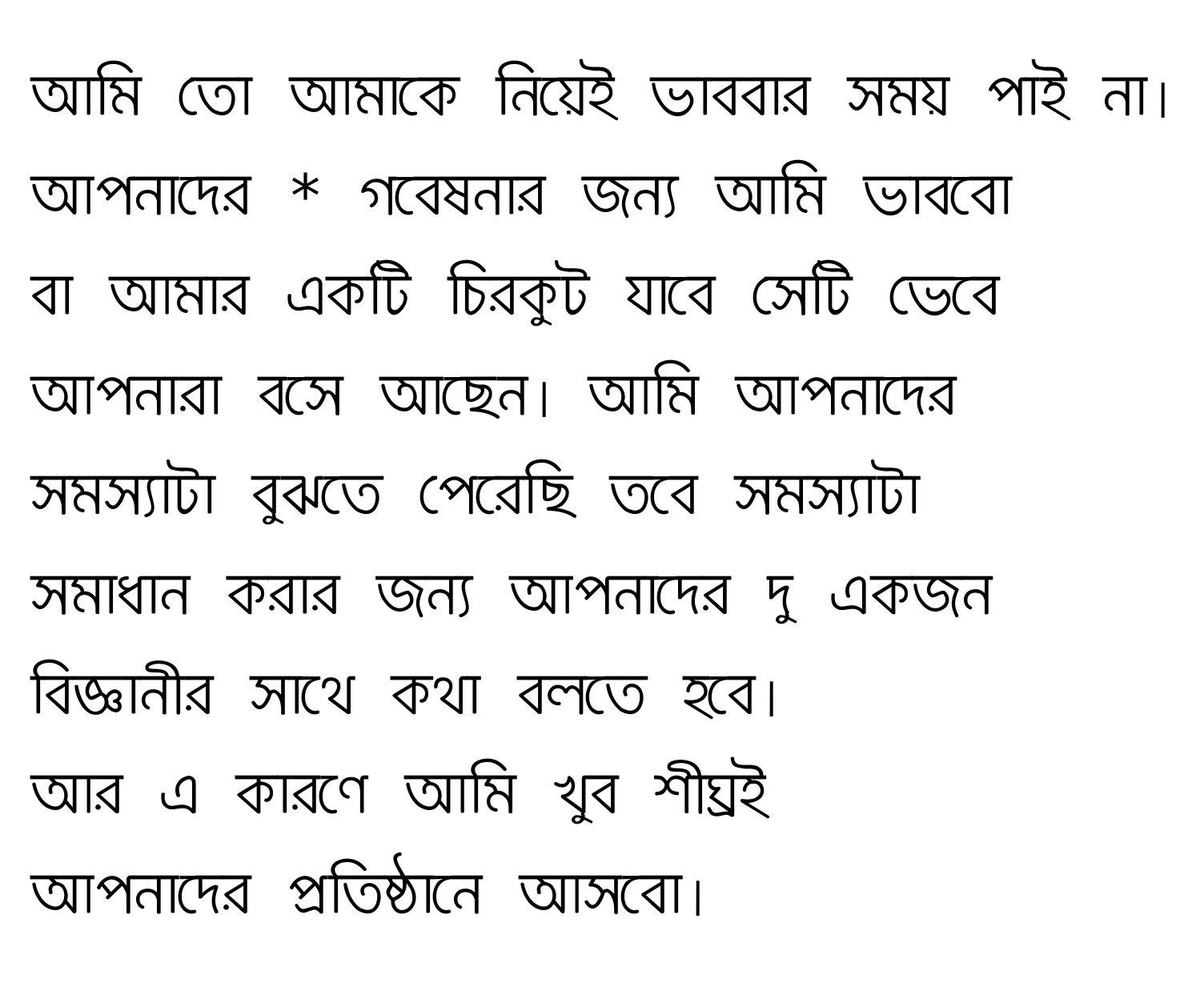}
		\caption{The left image illustrates a handwriting image with word-level bounding-boxes. The labels/words for each bounding-box is presented on the right. The excluded word (second row, second word) is marked using an asterisk (*).}
		\label{fig:labelview}
	\end{center}
\end{figure}

The BanglaWriting dataset contains single-page handwritings of 260 individuals from eight different districts (illustrated in Table \ref{tab:source_distribution}). It consists of 5,470 unique words and 124 unique characters. Moreover, the overall dataset comprises 21,234 words and 32,787 characters in total. The dataset contains Bangla characters, numerics, diacritics, and conjuncts. Furthermore, it has punctuation marks and English alphabets mixed with Bangla writing. Table \ref{tab:chars} illustrates the Bangla characters that exist in the dataset. For better understanding, Fig. \ref{fig:construction} explicates the underlying construction of a Bangla word. Fig. \ref{fig:labelview} illustrates a sample of the BanglaWriting dataset, bounding-box, and labels.

\begin{table}[!h]
	\centering
	\caption{The BanglaWriting dataset contains all characters of Bangla vocabulary. The table illustrates the Bangla characters that also exist in the dataset.}
	\label{tab:chars}
	\includegraphics[width=\linewidth]{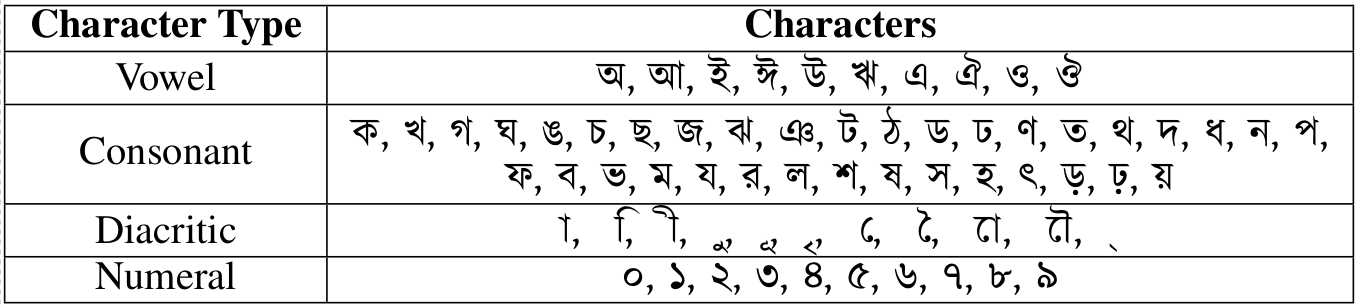}
\end{table}

\begin{figure}[!h]
	\centering
	\includegraphics[width=0.4\linewidth]{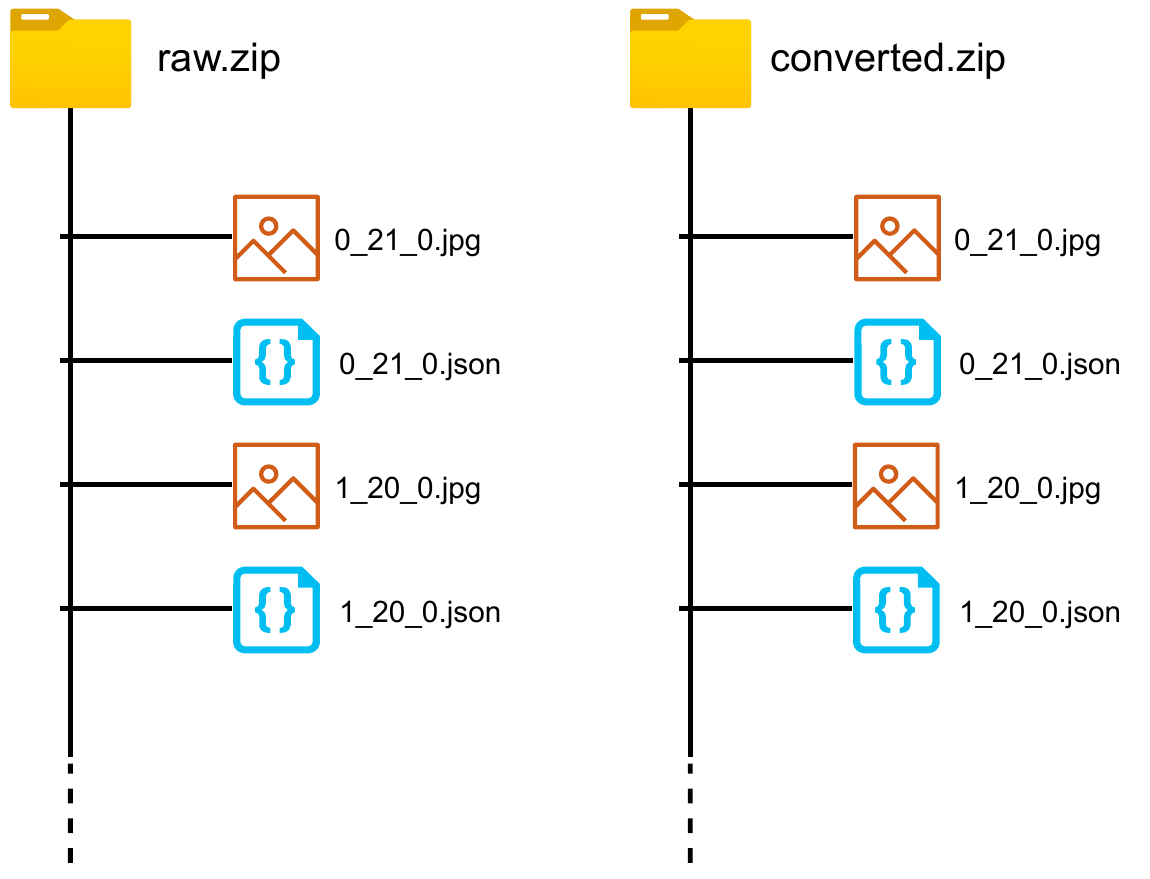}
	\caption{The figure illustrates the directory structure of the BanglaWriting data files. The 'raw.zip' contains raw images that were only labeled. The 'converted.zip' contains labels, and the images are manually processed using the additional script \cite{banglawriting_script2020}. For every image file, there exists a \emph{JSON} file with the same naming scheme. The \emph{JSON} file contains the bounding-boxes and labels.}
	\label{fig:dir}
\end{figure}

\begin{figure}[!h]
	\centering
	\includegraphics[height=0.8\textheight]{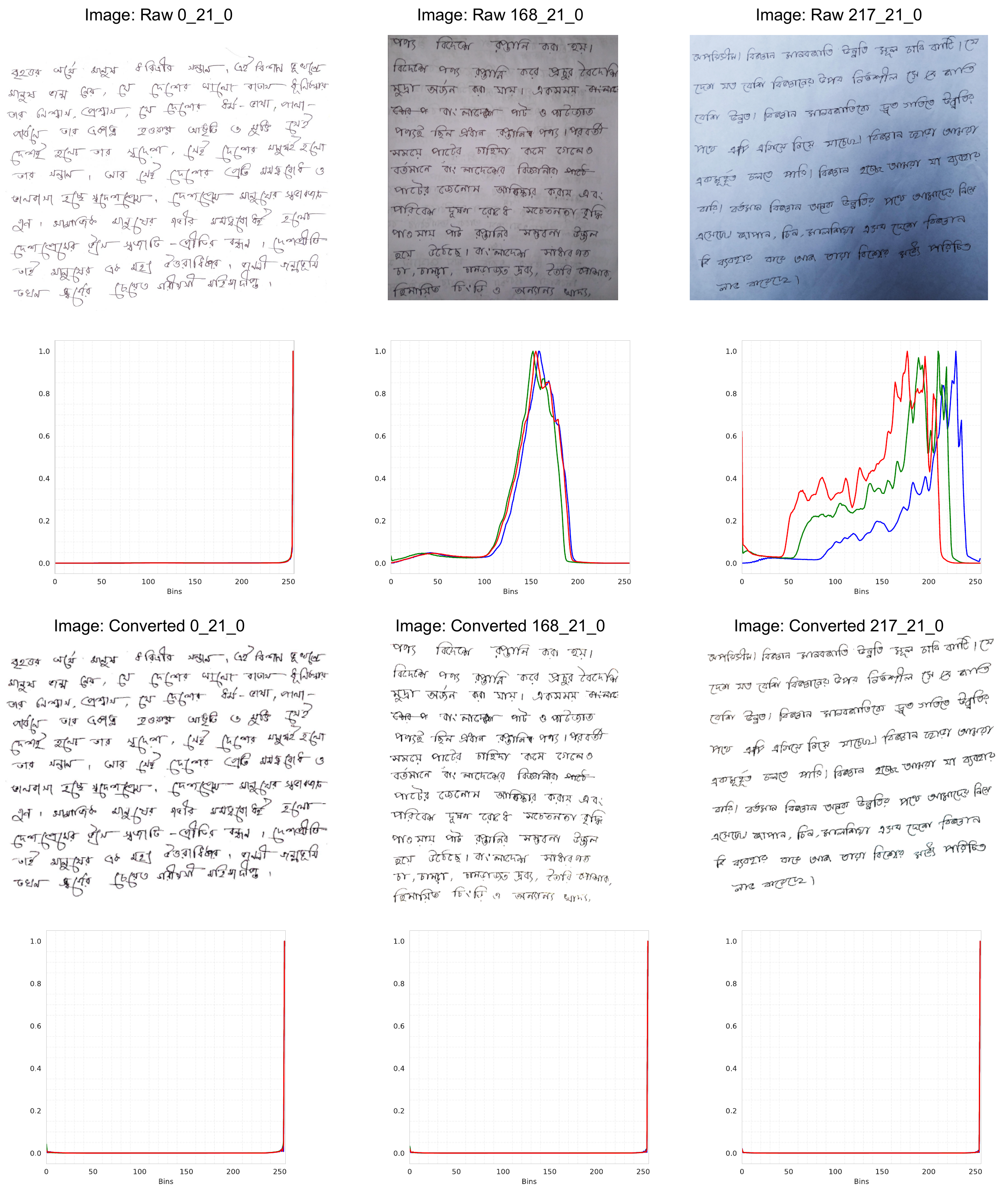}
	\caption{The illustration points out the image data variation in the 'raw' and 'converted' versions of the BanglaWriting dataset. The upper row illustrates the raw version's image data, where the first image is taken using a scanner, and the rest are captured using a smartphone camera. The second row illustrates color histograms w.r.t. the images. The third row depicts the same pictures from the converted version (processed using the supplementary script \cite{banglawriting_script2020}). The fourth row illustrates the color histogram w.r.t. the images in the third row. By comparing the color histograms, it can be concluded that the 'raw' version's images contain color shifts and light issues. In contrast, the converted images exclude those challenges.}
	\label{fig:raw_vs_conv}
\end{figure}

The dataset is presented in two different versions, i) raw and ii) converted. The raw file contains raw images that were manually cropped, and no image-processing techniques were applied. Hence, the raw dataset includes a diversity of color shifts, shadowing effects in images. On the contrary, the converted file contains a furnished version of the raw images (discussed in Section \ref{sec:supplementaryscript}). Fig. \ref{fig:raw_vs_conv} illustrates the difference between the raw and converted dataset images. Further, Fig. \ref{fig:dir} shows the directory structure for both dataset versions. 

Each of the pictures in both datasets comprises the writing of a single individual. Individual images are named based on the following convention, 

\begin{equation*} 
	personIdentifier\_age\_gender.jpg
\end{equation*}
\begin{equation*}
	\begin{split}
		Where, &\\
		&personIdentifier = \text{Unique id assigned to an individual.}\\
		&age =  \text{Age of the individual.}\\
		&gender = \text{Gender of the individual. 0 for females and 1 for males.}\\
	\end{split}
\end{equation*}

For every image data, a \emph{JSON} file is also included with the same naming convention. The \emph{JSON} file contains the word-level bounding-box information and labels for each bounding-box. The \emph{JSON} format is illustrated in Fig. \ref{fig:json} and it is further elaborated in Section \ref{sec:datalabeling}.

The labels for each word-level bounding-box represents the words written in unicode format. There are three possible classes/label-formats maintained, which are presented below. 


\begin{enumerate}
	\item \textbf{Clear writing}: By clear writing, we refer if the bounding-box contains written word that the writer intended to write and are understandable. In this case, we label the bounding-box with the unicode value of the written word.
	
	\item \textbf{Overwriting}: By overwriting, we refer if the bounding-box contains the written word, but some of the characters have been stroked out. Writers often strike-out some character to refer to exclude that character. In such a case, we label the comprehensible characters with proper unicodes, and we omit the stroked out characters in the label. In such a case, we add an asterisk ('*') with the Unicode label to mark the issue.
	
	\item \textbf{Strikes and mistakes}: The dataset contains some random strikes (such as word underlines, rules), and fully stroked out words. We do not include any unicode in such cases, and we only label them using an asterisk ('*').
\end{enumerate} 

\begin{figure}[!h]
	\centering
	\includegraphics[width=\linewidth]{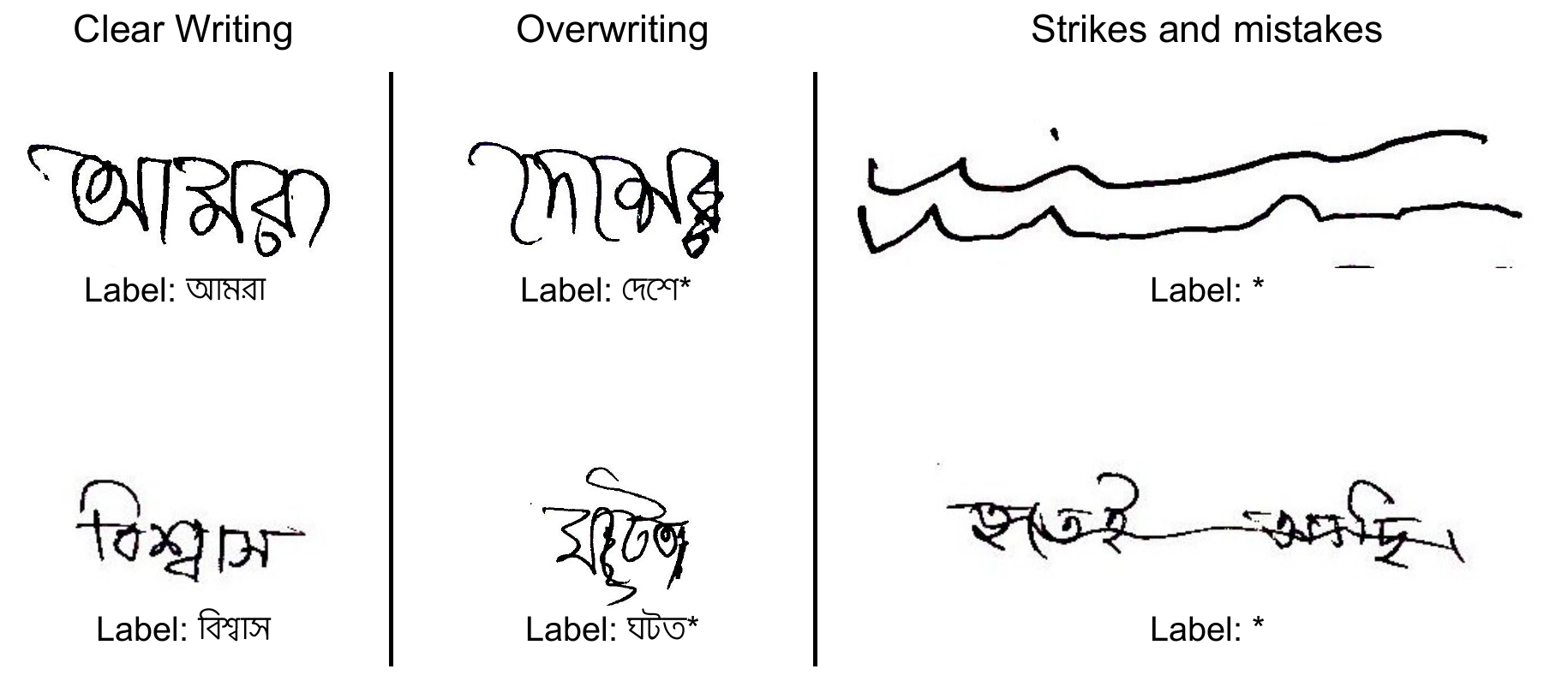}
	\caption{The figure depicts some examples of the words and labels generated for each class. The left, middle, and right columns explicate clear writing, overwriting, and strikes/mistakes, respectively.}
	\label{fig:label_example}
\end{figure}

Fig. \ref{fig:label_example} further illustrates some examples of the labels mentioned above. Moreover, Table \ref{tab:class_distribution} represents the quantitative distribution of each class in the dataset.

The dataset also includes a supplementary script \cite{banglawriting_script2020} used to produce the furnished images of the 'converted' version of the data. The script is used to reduce the noises and light variations of the 'raw' data images.

\begin{table}	
	\centering
	\caption{The table describes the quantitative distribution of each label along with the labeling schemes.}
	\label{tab:class_distribution}
	\begin{tabular}{|c|c|c|}
		\hline
		\textbf{Class} & \textbf{Count} & \textbf{Label Scheme} \\ \hline
		Clear writing & 21234 & Contains word in unicode \\ \hline
		Overwriting & 261 & Contains word in unicode with asterisk '*' \\ \hline
		Strike and mistakes & 450 & Contains asterisk '*' \\ \hline
	\end{tabular}
\end{table}

\begin{table}
	\centering
	\caption{The table describes the quantitative distribution of the geographical location of the writers.}
	\label{tab:source_distribution}
	\begin{tabular}{|c|c|}
		\hline
		\textbf{District} & \textbf{Total Documents} \\ \hline
		Dhaka & 48 \\ \hline
		Gopalganj & 26 \\ \hline
		Comilla & 14 \\ \hline
		Gazipur & 21 \\ \hline
		Tangail & 36 \\ \hline
		Netrakona & 25 \\ \hline
		Kishoreganj & 46 \\ \hline
		Mymensingh & 44 \\ \hline
	\end{tabular}
\end{table}

\section{Experimental Design, Materials and Methods}

\subsection{Data Collection}



\begin{figure}[!h]
	\begin{center}
		\includegraphics[width=\linewidth]{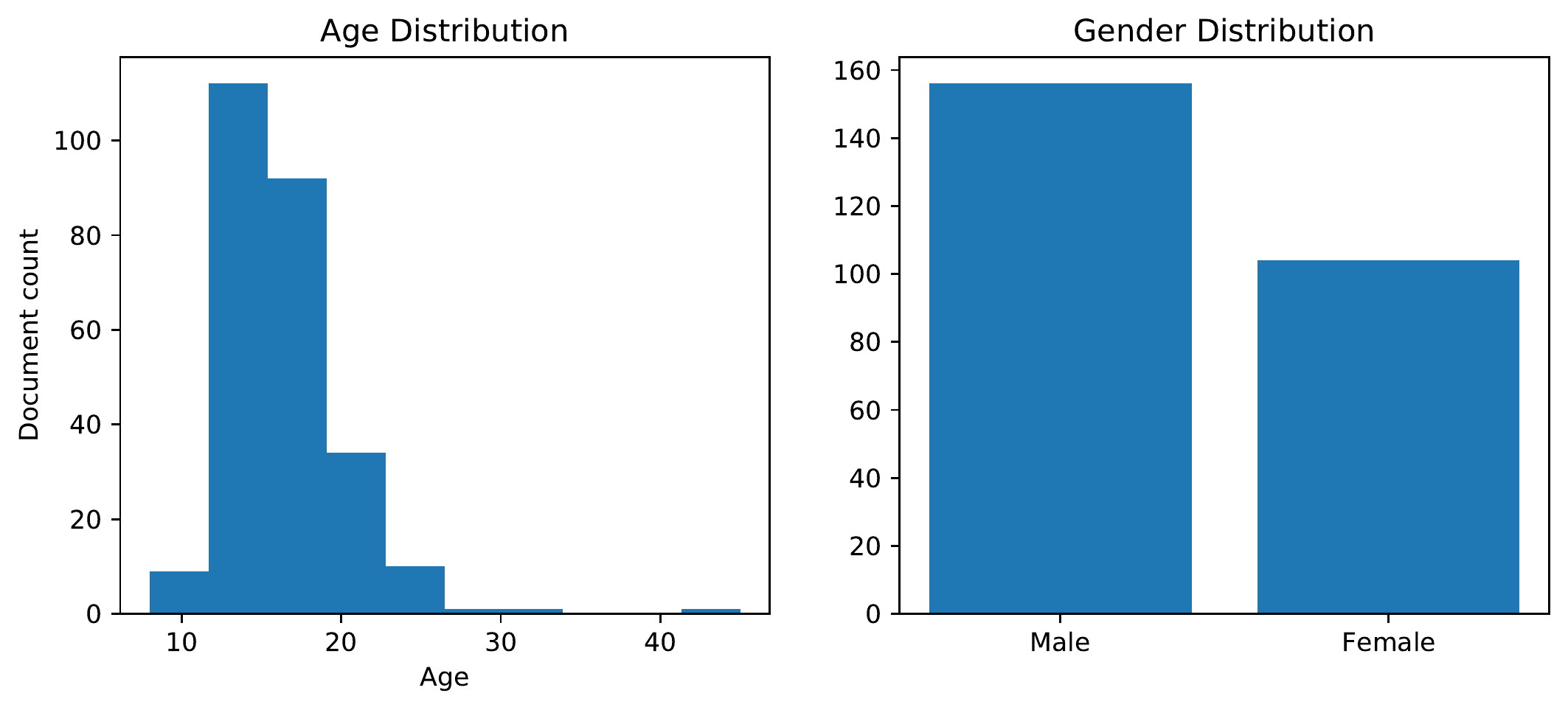}
		\caption{The left graph exhibits age distribution, and the right graph demonstrates the gender distribution of the dataset.}
		\label{fig:age_gender}
	\end{center}
\end{figure}

The dataset was collected from the students of Bangladesh University of Business and Technology. Furthermore, to generate a better age distribution of the dataset, the students' household members were also included. Fig. \ref{fig:age_gender} illustrates the age and gender distribution of the population. However, the writers were selected based on the primary clinical constraints, a) The minimum age of the writers can be 8, b) The writers should be physically fit to write.

\begin{figure}[!h]
	\begin{center}
		\includegraphics[width=\linewidth]{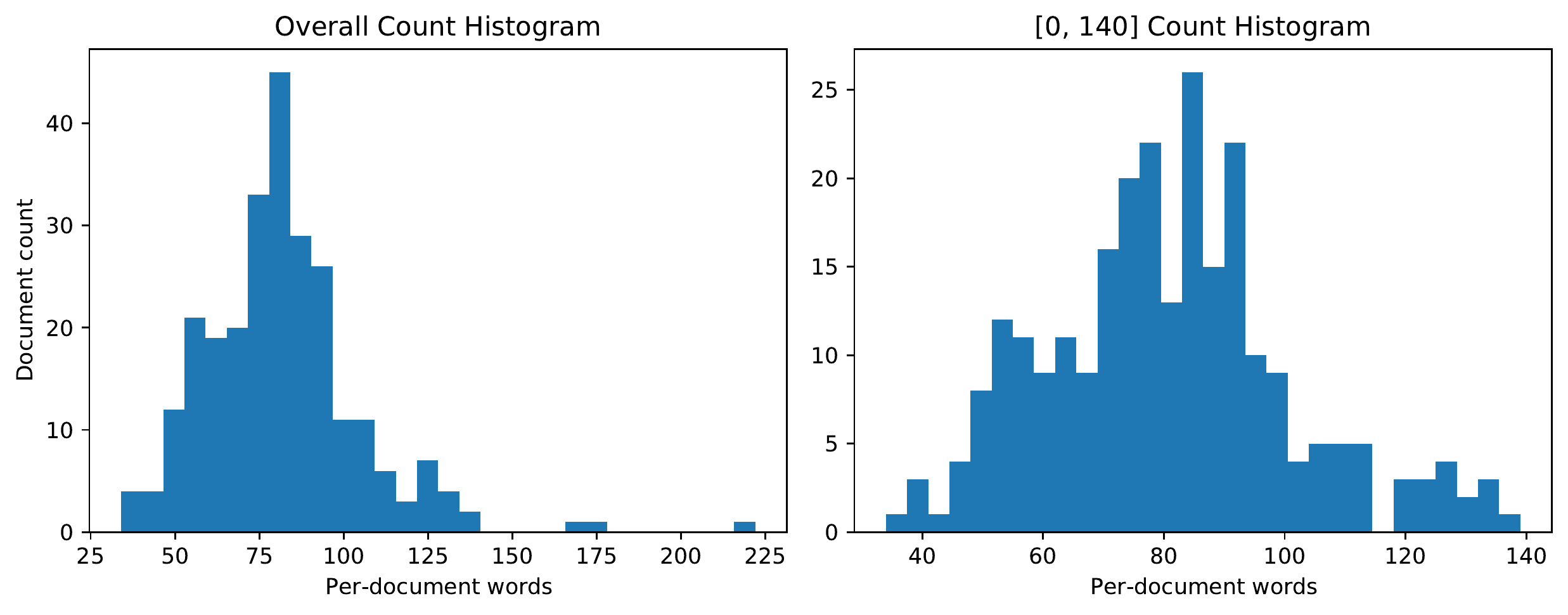}
		\caption{The left graph illustrates the word per document distribution for each paper. The right graph shows the same scenario without outliers. The word-count histogram simulates normal distribution.}
		\label{fig:wordcount}
	\end{center}
\end{figure}

The writers written on A4-sized papers, and regular ball-point and gel pens were used for writing. Each individual was suggested to write on any topic. Therefore, each document contains a diverse number of words. Fig. \ref{fig:wordcount} represents the word distribution per document. Moreover, allowing writers to write on random topics also resulted in making mistakes and overwriting that are also labeled.

The writers are from eight different districts of Bangladesh. We define a writer belonging to a particular district if he/she stayed in the district for more than ten years. Table \ref{tab:source_distribution} illustrates a quantitative distribution of the geographical location of the writers.

\subsection{Data Extraction}
\label{sec:dataextraction}

The handwritten pages were further imaged using a scanner and smartphone cameras. The dataset contains a total of 52 scanned images and 208 images captured using smartphone cameras. The scanned images do not contain any noisy conditions. On the contrary, the images captured using smartphone cameras have noises due to environmental factors, such as various lighting effects, glazes of flashlight, and shadow effects.

\subsection{Data Preprocessing}
\label{sec:datapreprocessing}

Each image data were cropped and strengthened manually. The images were named using the formula, $personIdentifier\_age\_gender$. No augmentation was applied to increase the dataset's size to ensure the dataset's authenticity and quality.

\begin{figure}[!h]
	\begin{lstlisting}[language=json,firstnumber=1]
{
    "shapes": [
    {"label": "wordLabel",
        "points": [[xmin, ymin], [xmax, ymax]]
    },
    {"label": "wordLabel",
        "points": [[xmin, ymin], [xmax, ymax]]
    },
    ....
    ],
    "imagePath": "uniquePersonIdentifier_age_gender.jpg",
    "imageHeight": Xpx,
    "imageWidth": Ypx
}
	\end{lstlisting}
	\caption{The figure illustrates a JSON structure that interprets the bounding-boxes and labels information for each handwriting image data.}
	\label{fig:json}
\end{figure}

\subsection{Data Labeling}
\label{sec:datalabeling}

The dataset was manually annotated using \emph{labelme} \cite{labelme2016} software. Fig. \ref{fig:labelview} illustrates the word-based bounding-boxes and the unicode-text labels for each bounding-box. The figure also demonstrates the annotation policy adapted for overwriting and cropped words/characters. Table \ref{tab:class_distribution} illustrates the labeling policy adopted for three different labels/classes of the word-based bounding-boxes.

The bounding-box and label information for each image was separately saved on individual JSON files, following the same naming convention of the handwritten images. Fig. \ref{fig:json} illustrates the standard JSON-file parameters that were generated for each image. The \emph{"shape"} property contains an array of \emph{"label"} and \emph{"points"} parameter pairs. The \emph{"label"} parameter contains the written word (in unicode-8) in the bounding-box. Whereas, the \emph{"points"} parameter contains an array of starting and ending pixel-coordinates of the bounding-box. The \emph{"imagePath"}, \emph{"imageHeight"}, and \emph{"imageWidth"} contains some additional information such as, the filename of the corresponding image, the height and width of the image, respectively.

\subsection{Supplementary Script}
\label{sec:supplementaryscript}

As the dataset contains raw images taken using scanners and smartphones, a difference of lightning and background noise is noticed (illustrated in Fig. \ref{fig:raw_vs_conv}). Hence, the dataset includes a supplementary \emph{Python} \cite{rossum1995python} and \emph{OpenCV} \cite{opencv_library} based script \cite{banglawriting_script2020} that eliminates lightning issues and reduces the background noises. The script further furnishes the images and generates images suitable for machine learning and deep learning strategies. The furnished images are provided in the 'converted.zip' file, whereas the 'raw.zip' contains the raw images where no image-processing techniques were applied.

\clearpage

\section*{Ethics Statement}
All the handwritings were obtained with the consent of the individuals who had participated in the writing.

\section*{Acknowledgments}
The authors would like to thank the Advanced Machine Learning (AML) lab and the Bangladesh University of Business and Technology (BUBT) for their resource sharing and precious suggestions.

\section*{Declaration of Competing Interest}
The authors declare that they have no known competing financial interests or personal relationships which have, or could be perceived to have, influenced the work reported in this article. 

\bibliographystyle{unsrt}
\bibliography{refs.bib}

\begin{thebibliography}{10}

\bibitem{labelme2016}
Kentaro Wada.
\newblock {labelme: Image Polygonal Annotation with Python}.
\newblock \url{https://github.com/wkentaro/labelme}, 2016.

\bibitem{michie1994machine}
Donald Michie, David~J Spiegelhalter, CC~Taylor, et~al.
\newblock Machine learning.
\newblock {\em Neural and Statistical Classification}, 13(1994):1--298, 1994.

\bibitem{lecun2015deep}
Yann LeCun, Yoshua Bengio, and Geoffrey Hinton.
\newblock Deep learning.
\newblock {\em nature}, 521(7553):436--444, 2015.

\bibitem{ohi2020autoembedder}
Abu~Quwsar Ohi, MF~Mridha, Farisa~Benta Safir, Md~Abdul Hamid, and
  Muhammad~Mostafa Monowar.
\newblock Autoembedder: A semi-supervised dnn embedding system for clustering.
\newblock {\em Knowledge-Based Systems}, 204:106190, 2020.

\bibitem{marti2002iam}
U-V Marti and Horst Bunke.
\newblock The iam-database: an english sentence database for offline
  handwriting recognition.
\newblock {\em International Journal on Document Analysis and Recognition},
  5(1):39--46, 2002.

\bibitem{biswas2017banglalekha}
Mithun Biswas, Rafiqul Islam, Gautam~Kumar Shom, Md~Shopon, Nabeel Mohammed,
  Sifat Momen, and Anowarul Abedin.
\newblock Banglalekha-isolated: A multi-purpose comprehensive dataset of
  handwritten bangla isolated characters.
\newblock {\em Data in brief}, 12:103--107, 2017.

\bibitem{mahmoud2014khatt}
Sabri~A Mahmoud, Irfan Ahmad, Wasfi~G Al-Khatib, Mohammad Alshayeb,
  Mohammad~Tanvir Parvez, Volker M{\"a}rgner, and Gernot~A Fink.
\newblock Khatt: An open arabic offline handwritten text database.
\newblock {\em Pattern Recognition}, 47(3):1096--1112, 2014.

\bibitem{grosicki12008rimes}
Emmanu{\`e}le Grosicki$^1$, Matthieu Carre, Jean-Marie Brodin, and Edouard
  Geoffrois$^1$.
\newblock Rimes evaluation campaign for handwritten mail processing.
\newblock 2008.

\bibitem{schomaker2000forensic}
Lambert Schomaker and Louis Vuurpijl.
\newblock {\em Forensic writer identification: A benchmark data set and a
  comparison of two systems}.
\newblock NICI (NIjmegen Institute of Cognitive Information), Katholieke
  Universiteit~…, 2000.

\bibitem{al2002data}
Somaya Al-Ma'adeed, Dave Elliman, and Colin~A Higgins.
\newblock A data base for arabic handwritten text recognition research.
\newblock In {\em Proceedings eighth international workshop on frontiers in
  handwriting recognition}, pages 485--489. IEEE, 2002.

\bibitem{banglaAI_2019}
Kaggle.
\newblock {Bengali.AI Handwritten Grapheme Classification}.
\newblock
  \url{https://www.kaggle.com/competitions/bengaliai-cv19/discussion/122421#699069},
  2019.

\bibitem{alam2021large}
Samiul Alam, Tahsin Reasat, Asif~Shahriyar Sushmit, Sadi~Mohammad Siddique,
  Fuad Rahman, Mahady Hasan, and Ahmed~Imtiaz Humayun.
\newblock A large multi-target dataset of common bengali handwritten graphemes.
\newblock In {\em International Conference on Document Analysis and
  Recognition}, pages 383--398. Springer, 2021.

\bibitem{banglawriting_script2020}
Abu~Quwsar Ohi.
\newblock {BanglaWriting: A multi-purpose offline Bangla handwriting dataset
  (script)}.
\newblock \url{https://github.com/QuwsarOhi/BanglaWriting}, 2020.

\bibitem{rossum1995python}
Guido Rossum.
\newblock Python reference manual.
\newblock 1995.

\bibitem{opencv_library}
G.~Bradski.
\newblock {The OpenCV Library}.
\newblock {\em Dr. Dobb's Journal of Software Tools}, 2000.

\end{thebibliography}


\begin{thebibliography}{1}

\bibitem{biswas2017banglalekha}
Mithun Biswas, Rafiqul Islam, Gautam~Kumar Shom, Md~Shopon, Nabeel Mohammed,
  Sifat Momen, and Anowarul Abedin.
\newblock Banglalekha-isolated: A multi-purpose comprehensive dataset of
  handwritten bangla isolated characters.
\newblock {\em Data in brief}, 12:103--107, 2017.

\bibitem{marti2002iam}
U-V Marti and Horst Bunke.
\newblock The iam-database: an english sentence database for offline
  handwriting recognition.
\newblock {\em International Journal on Document Analysis and Recognition},
  5(1):39--46, 2002.

\bibitem{michie1994machine}
Donald Michie, David~J Spiegelhalter, CC~Taylor, et~al.
\newblock Machine learning.
\newblock {\em Neural and Statistical Classification}, 13(1994):1--298, 1994.

\bibitem{lecun2015deep}
Yann LeCun, Yoshua Bengio, and Geoffrey Hinton.
\newblock Deep learning.
\newblock {\em nature}, 521(7553):436--444, 2015.

\bibitem{ohi2020autoembedder}
Abu~Quwsar Ohi, MF~Mridha, Farisa~Benta Safir, Md~Abdul Hamid, and
  Muhammad~Mostafa Monowar.
\newblock Autoembedder: A semi-supervised dnn embedding system for clustering.
\newblock {\em Knowledge-Based Systems}, 204:106190, 2020.

\bibitem{labelme2016}
Kentaro Wada.
\newblock {labelme: Image Polygonal Annotation with Python}.
\newblock \url{https://github.com/wkentaro/labelme}, 2016.

\bibitem{rossum1995python}
Guido Rossum.
\newblock Python reference manual.
\newblock 1995.

\bibitem{opencv_library}
G.~Bradski.
\newblock {The OpenCV Library}.
\newblock {\em Dr. Dobb's Journal of Software Tools}, 2000.

\bibitem{banglawriting_script2020}
Abu~Quwsar Ohi.
\newblock {BanglaWriting: A multi-purpose offline Bangla handwriting dataset
  (script)}.
\newblock \url{https://github.com/QuwsarOhi/BanglaWriting}, 2020.

\end{thebibliography}
\end{document}